\documentclass{article}

\usepackage{multirow}
\usepackage{subcaption}
\usepackage{caption}
\usepackage{spconf,amsmath,graphicx}
\usepackage{amsfonts}

\usepackage{xcolor}
\usepackage{wrapfig}
\usepackage{cite}
\usepackage{booktabs}

\usepackage{threeparttable}
\usepackage{colortbl}
\usepackage{setspace}
\usepackage{enumitem}
\usepackage{cite}
\usepackage{booktabs}
\usepackage{threeparttable}
\usepackage{makecell}
\usepackage{hyperref}
\hypersetup{hypertex=true,
            colorlinks=true,
            linkcolor=black,
            anchorcolor=black,
            citecolor=black}

\usepackage{threeparttable}
\usepackage{pdfpages}
\usepackage{tabularx}
\usepackage{soul}
\usepackage[font=small,labelsep=none]{caption}

\title{ESIHGNN: Event-State Interactions Infused Heterogeneous Graph Neural Network for Conversational Emotion Recognition}

\name{Xupeng Zha \qquad Huan Zhao$^{\star}$ \qquad Zixing Zhang$^{\star}$
\thanks{$^\star$ Corresponding authors: H. Zhao and Z. Zhang (email: \{hzhao, zixingzhang\}@hnu.edu.cn). The work was funded by the National Science Foundation of China under Grant Number 62076092.}
}

\address{College of Computer Science and Electronic Engineering, Hunan University, China}

\usepackage[switch]{lineno}
 
\begin{document}
\ninept
\maketitle
%

\begin{abstract}
Conversational Emotion Recognition (CER) aims to predict the emotion expressed by an utterance (referred to as an ``event") during a conversation.
Existing graph-based methods mainly focus on event interactions to comprehend the conversational context, while overlooking the direct influence of the speaker's emotional state on the events.
In addition, real-time modeling of the conversation is crucial for real-world applications but is rarely considered.
Toward this end, we propose a novel graph-based approach, namely Event-State Interactions infused Heterogeneous Graph Neural Network (ESIHGNN), which incorporates the speaker's emotional state and constructs a heterogeneous event-state interaction graph to model the conversation.
Specifically, a heterogeneous directed acyclic graph neural network is employed to dynamically update and enhance the representations of events and emotional states at each turn, thereby improving conversational coherence and consistency.
Furthermore, to further improve the performance of CER, we enrich the graph's edges with external knowledge.
Experimental results on four publicly available CER datasets show the superiority of our approach and the effectiveness of the introduced heterogeneous event-state interaction graph.

\end{abstract}

\vspace{-.02cm}
\section{Introduction}
\vspace{-.02cm}
Evidence from psychology suggests that human actions are influenced by numerous factors, including environmental events and emotional ambiance~\cite{psychology:1, psychology:2}.
Understanding the decision-making process underlying human actions is crucial, as it can reflect individuals' subjective evaluations of these factors.
Recognizing emotions in conversations is an important foundation for this endeavor~\cite{conversation-emotion:1, conversation-emotion:2}.

Conversational Emotion Recognition (CER) is a crucial cognitive task that aims to identify the emotions conveyed by speakers through their utterances (referred to as ``events'') during a conversation.
CER has attracted significant attention in multiple research fields, as it serves as the foundation for developing conversational agents with emotional intelligence.
These agents find applications in domains such as virtual reality therapy~\cite{virtual-reality-therapy}, social robotics~\cite{social-robotics}, and smart home systems~\cite{smart-home-systems}.
In reality, the flow of a conversation is influenced by both previous events and the emotional states of the participants, which can trigger new events and update their emotions.
Therefore, accurately modeling the real-time conversation driven by previous events and emotional states is crucial for a successful CER.

Recent studies have focused on integrating speaker information into conversation modeling.
Two approaches have been researched: recurrence-based methods and graph-based methods.
Recurrence-
based methods encode an event flow and a parallel state flow, and build interactions between them to dynamically model the conversation.
For example, DialogueRNN~\cite{DialogueRNN} tracks the state of each speaker along the event flow using different Gated Recurrent Units (GRUs).
However, it has limitations in terms of scalability and capacity to efficiently capture the interactions between events and speakers' states.
To tackle scalability, DialogueCRN~\cite{DialogueCRN}, CoMPM~\cite{CoMPM}, and DialogueINAB~\cite{DialogueINAB} propose a unified speaker-level emotion tracking module.
To enhance natural interactions between events and states, COSMIC~\cite{COSMIC} extends DialogueRNN with external knowledge.
However, it still remains challenging to balance long- and short-term memory within each utterance.
Moreover, these methods often require bidirectional context modeling, which sacrifices real-time conversational abilities while improving performance.

On the other hand, graph-based methods construct a conversation structure graph constrained by speaker identity to model the conversation using Graph Neural Networks~(GNNs).
DialogueGCN~\cite{DialogueGCN}, for example, represents events as nodes and connections between speakers as edges to gather contextual information from neighboring nodes within a specific window.
RGAT~\cite{position-RGAT} expands upon DialogueGCN by including position encodings to consider sequential information.
Meanwhile, DAG-ERC~\cite{DAG-ERC} constructs a directed acyclic graph inspired by the Directed Acyclic Graph Neural Network~(DAGNN)~\cite{DAGNN} to sequentially model context.
Similar to COSMIC, SKAIG~\cite{SKAIG} enhances event interactions with external knowledge, while knowledge also plays an essential role in both KET~\cite{KET} and KI-Net~\cite{KI-Net}.
In summary, graph-based methods effectively model conversations by constructing interaction networks.
However, these methods overlook the emotional states of speakers during a conversation and focus only on event interactions.

According to the aforementioned observations, it is important to develop an event-state interaction graph-based approach that can capture effective interactions between events and the emotions of participants for real-time conversation modeling.
This paper proposes a solution named \textbf{ESIHGNN}~(\textbf{E}vent-\textbf{S}tate \textbf{I}nteractions infused \textbf{H}eterogeneous \textbf{G}raph \textbf{N}eural \textbf{N}etwork) to meet these requirements.
Specifically, we construct a heterogeneous event-state graph, where the event node is initialized with the semantic feature of the utterance and the accompanying state node representing the speaker's emotional state is initialized with the speaker identity.
To support real-time conversation modeling and enhance natural interactions between the event nodes and state nodes, we define eight preliminary logical edge relations based on speaker identity, represented by external knowledge or trainable vectors.
The resulting graph is then fed into an introduced \textbf{HDAGNN}~(\textbf{H}eterogeneous \textbf{D}irected \textbf{A}cyclic \textbf{G}raph \textbf{N}eural \textbf{N}etwork) that recurrently updates each event and state node in a single layer by gathering information from previous events and emotional states.
Additionally, we use GRUs to enhance the representations of event and state nodes at each turn, in order to improve conversational consistency.
Our contributions are summarized as follows:
(1) We propose an ESIHGNN, a novel approach that first treats a conversation as a heterogeneous event-state interaction graph for the CER task.
(2) We introduce an HDAGNN, a heterogeneous directed acyclic graph neural network that dynamically models interactions between events and speakers' emotional states during real-time conversations.
(3) Extensive experiments conducted on four benchmark datasets validate the effectiveness and superiority of the proposed ESIHGNN and confirm the advantages of the heterogeneous event-state interaction graph.

\begin{figure*}[htb] 
\centering 
\includegraphics[width=0.88\linewidth]{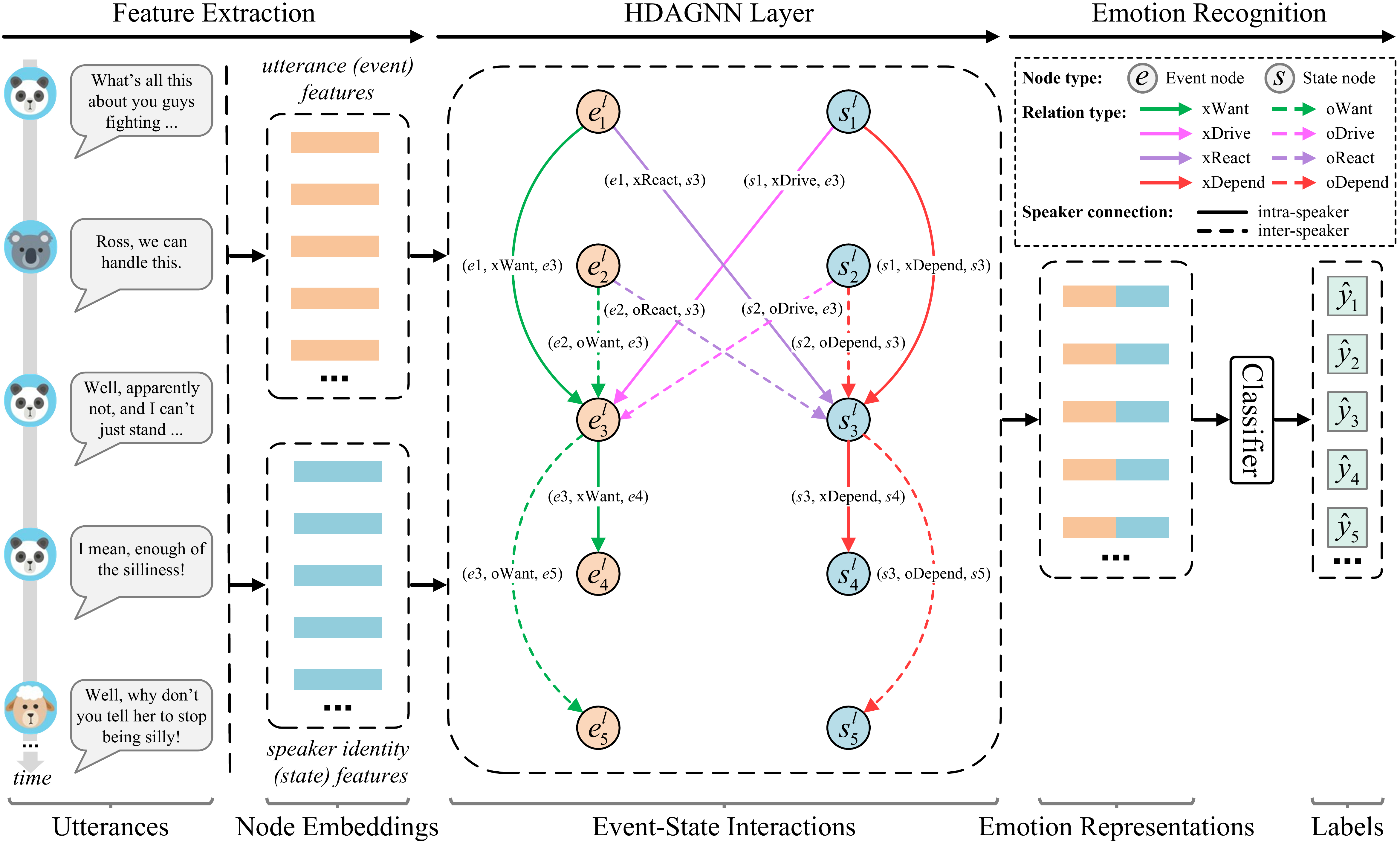} 
\vspace{-.1cm}
\caption{The introduced ESIHGNN framework, where we present the interactions between nodes of the 3rd turn and their predecessors and successors in a conversation example.
} 
\label{figure_1} 
\vspace{-.25cm}
\end{figure*}

\vspace{-.07cm}
\section{Methodology}
\vspace{-.07cm}
Our ESIHGNN comprises three primary modules:
graph construction, HDAGNN for feature transformation, and emotion prediction.
Figure~\ref{figure_1} illustrates the ESIHGNN framework.

\vspace{-.07cm}
\subsection{Task Definition}
\vspace{-.07cm}
In a conversation, a series of $N$ sequential utterances are encountered, denoted as $[(u_1, p_{u_1}),(u_2, p_{u_2}), \ldots,(u_N, p_{u_N})]$.
Here, $u_i$ represents the utterance of the $i$-th turn, and $p_{u_i}$ indicates the speaker identity for $u_i$.
Note that $p_{u_i}$ and $p_{u_j}$ may belong to the same speaker, but there must be at least two speakers participating in the conversation.
The task of the CER is to predict the emotion category for each utterance.
For a target utterance $u_i$, its prediction relies on real-time sequence pairs consisting of utterances and accompanying speaker identities: $\left\{(u_1, p_{u_1}), (u_2, p_{u_2}), \ldots, (u_i, p_{u_i})\right\}$.

\vspace{-.07cm}
\subsection{Graph Construction}
\vspace{-.07cm}
\label{sec:Graph construction}
A new methodology for modeling conversation structure is presented, emphasizing the interactions between events and speakers' emotional states.
This methodology constructs a heterogeneous event-state interaction graph, denoted as $\mathcal{G}=(\mathcal{V}, \mathcal{E}, \mathcal{R}, \mathcal{A})$, where $\mathcal{V}$ represents the node set consisting of event nodes $\mathcal{V}_e$ and state nodes $\mathcal{V}_s$.
In this graph, an edge $e_{ij} = \left(v_i, r, v_j\right) \in \mathcal{E}$ signifies a connection from a node $v_i \in \mathcal{V}$ to its neighboring node $v_j \in \mathcal{V}$ under relation type $r \in \mathcal{R}$, with its representation denoted by $a_{ij,r} \in \mathcal{A}$.

\noindent
\textbf{Nodes:}
For sequential utterances in a conversation, each utterance $u_i$ is regarded as an event node $v_{e_i}$, while the emotional state of $u_i$' speaker is represented as a state node $v_{s_i}$.
Clearly, both $v_{e_i}$ and $v_{s_i}$ correspond to the same speaker and turn.
Consistent with COSMIC~\cite{COSMIC} and SKAIG~\cite{SKAIG}, we employ the fine-tuned RoBERTa model~\cite{RoBERTa} to encode the utterance $u_i$ as the initial event node feature $h_{e_i}$.
The state node feature $h_{s_i}$ is initialized using a one-hot vector that indicates the speaker identity $p_{u_i}$.
The context-independence of these node features is essential for modeling real-time conversations.

\noindent
\textbf{Relations:}
To simulate a real-time conversation driven by previous events and speakers' emotional states, we suggest eight preliminary event-state interaction relations to capture the historical context.

\vspace{0.1cm}
\hspace{-0.15cm}\textbf{Event Node Updating:}
\begin{itemize}[leftmargin=0.66cm,itemsep=-.2pt]
\item \texttt{xWant}: Intra-speaker $event\texttt{-}to\texttt{-}event$ interaction:
Event $e_i$ passes the speaker's action guidance to the subsequent event $e_j$.

\item \texttt{oWant}: Inter-speaker $event\texttt{-}to\texttt{-}event$ interaction:
Event $e_i$ can coordinate and motivate the subsequent event $e_j$ triggered by another speaker.

\item \texttt{xDrive}: Intra-speaker $state\texttt{-}to\texttt{-}event$ interaction:
The speaker's emotional state $s_i$ internally influences the trend of their subsequent event $e_j$.

\item \texttt{oDrive}: Inter-speaker $state\texttt{-}to\texttt{-}event$ interaction:
The speaker's emotional state $s_i$, resulting from contagion and empathy, externally influences the trend of the subsequent event $e_j$ triggered by another speaker.

\end{itemize}

\hspace{-0.15cm}\textbf{State Node Updating:}
\begin{itemize}[leftmargin=0.66cm,itemsep=-.2pt]

\item \texttt{xReact}: Intra-speaker $event\texttt{-}to\texttt{-}state$ interaction:
When event $e_i$ is executed by the speaker, their subsequent state $s_j$ is influenced by their reaction to this event $e_i$.

\item \texttt{oReact}: Inter-speaker $event\texttt{-}to\texttt{-}state$ interaction:
The reaction of another speaker to an event $e_i$ influences their subsequent state $s_j$.

\item \texttt{xDepend}: Intra-speaker $state\texttt{-}to\texttt{-}state$ interaction:
The speaker's emotional state $s_i$ at a given moment has an influence on their subsequent emotional state $s_j$, indicating self-dependency.

\item \texttt{oDepend}: Inter-speaker $state\texttt{-}to\texttt{-}state$ interaction:
The emotional states of different speakers influence and interact with each other, indicating inter-speaker dependency.

\end{itemize}

\noindent
\textbf{Edges:}
Given the set of relation types, $\mathcal{R}$~=~\{\texttt{xWant}, \texttt{oWant}, \texttt{xDrive}, \texttt{oDrive}, \texttt{xReact}, \texttt{oReact}, \texttt{xDepend}, \texttt{oDepend}\}, information can only flow from previous event/state nodes to the current ones along the edges~$\mathcal{E}$, and not in the opposite direction.
To study the effect of the node connection range on the performance of CER, we introduce a window parameter $\omega$ that restricts the connections between the target node and previous nodes of each speaker within a window of size $\omega$.
With $\omega=1$, only the most recent event and state nodes are considered predecessors for the target node.

\noindent
\textbf{Edge Representations:}
External knowledge plays a crucial role in modeling conversations with fluency, coherence, and emotional contagion.
To harness this potential, we employ COMET~\cite{COMET}, an inferential, knowledge-based transformer model, to generate edge representations based on the input format from COSMIC~\cite{COSMIC} and SKAIG~\cite{SKAIG}.
For each edge $e_{ij} = \left(v_i, r, v_j\right)$, COMET takes the concatenated predecessor $v_i$ and relation $r$ as input, and extracts the hidden state of the relation token to serve as the edge representation $a_{ij,r}$.
Note that COMET was trained on commonsense knowledge data consisting of explicit event prompts, enabling it to generate edge representations for predecessor event nodes.
However, COMET does not generate edge representations for predecessor state nodes associated with implicit header entities, including \texttt{xDrive}, \texttt{oDrive}, \texttt{xDepend}, and \texttt{oDepend}.
To address this limitation, we propose using 300-dimensional trainable random vectors, which allow the model to learn these implicit relations during training.

\subsection{HDAGNN Layers}
We will now describe the methodology for feature transformation using HDAGNN.
In addition to establishing event-state interactions across different turns for conversational coherence (see Subsection \ref{sec:Graph construction}), it is equally crucial to consider the event-state interactions within the same turn to improve conversational consistency.
To achieve this, we propose establishing two implicit, bidirectional paths connecting the event and its accompanying emotional state.

\noindent
\textbf{Inter-Turn Event-State Interactions:}
To model real-time conversations, we use a dynamic and forward feature propagation strategy that enables current nodes to only receive information from previous nodes.
For each node $v_i$, the weight coefficients are calculated by normalizing the attention scores between the hidden feature of $v_i$ at the ($l\!-\!1$)-th layer and those of its neighborhood $\mathcal{N}_i$ at the $l$-th layer:
\begin{equation}
\label{equ1}
\alpha_{ij}^{l}=\mathop{\operatorname{Softmax}}\limits_{j\in\mathcal{N}_i, r\in\mathcal{R}}(W^{l}[W_{r,v}^{l}h_i^{l-1}+W_{r,v}^{l}h_j^l+W_{r,a}^{l}a_{ji,r}+W_{p}^{l}p_{j,r}]),
\end{equation}
where $W^{l}$, $W_{r,v}^{l}$, $W_{r,a}^{l}$, and $W_{p}^{l}$ are trainable parameters, and $p_{j,r}$ signifies the positional information of node $v_j$ using two dimensions.
The first dimension of $p_{j,r}$ indicates the absolute position, providing global sequential information throughout a conversation, while the second dimension denotes the relative position within the relation $r$, giving local sequential information within that relation.

Once obtained, the normalized coefficients are used to compute a linear combination of the features of $v_i$'s neighborhood $\mathcal{N}_i$, producing aggregated information (referred to as ``message''):
\begin{equation}
\label{equ2}
M_i^{l}=\sum\nolimits_{j\in{\cal N}_i}\alpha_{ij}^{l}W_{r,v}^{l}h_j^{l}.
\end{equation}

In HDAGNN, two GRUs are used to address the node heterogeneity and capture the sequential dependencies among nodes in an event-state interaction graph.
The GRU takes the past feature $h_i^{l-1}$ of the node $v_i$ as its input and the message $M_i^l$ as its hidden state.
\begin{equation}
\label{equ3}
\bar{h}_{e_i}^l={\text{GRU}}_{e,inter}^{l}(h_{e_i}^{l-1},M_{e_i}^{l}),
\end{equation}
\begin{equation}
\label{equ4}
\bar{h}_{s_i}^l={\text{GRU}}_{s,inter}^{l}(h_{s_i}^{l-1},M_{s_i}^{l}).
\end{equation}

\noindent
\textbf{Intra-Turn Event-State Interactions:}
Within a turn, there are two event-state interaction paths.
The first path involves the speaker being influenced by the emotional context when executing events.
For the emotional context of the $i$-th turn, we specify $M_{s_i}^{l}$ instead of the updated $\bar{h}_{s_i}^l$ that may have forgotten some emotional information.
\begin{equation}
\label{equ5}
\tilde{h}_{e_i}^l={\text{GRU}}_{e,intra}^{l}(M_{s_i}^{l},h_{e_i}^{l-1}),
\end{equation}
where $M_s^{l}$, $h_e^{l-1}$, and $\tilde{h}_e^l$ are the input, hidden state, and output of the GRU, respectively.

The second path involves the emotional state of the speaker at the $i$-th turn being influenced by the context of the event, $M_{e_i}^{l}$:
\begin{equation}
\label{equ6}
\tilde{h}_{s_i}^l={\text{GRU}}_{s,intra}^{l}(M_{e_i}^{l},h_{s_i}^{l-1}).
\end{equation}

To ensure conversational coherence and consistency, the hidden feature of the node $v_i$ at the $l$-th layer is the addition of $\bar{h}_i^l$ and $\tilde{h}_i^l$:
\begin{equation}
\label{equ7}
{h}_i^l=\bar{h}_i^l+\tilde{h}_i^l.
\end{equation}

\vspace{-.5cm}
\subsection{Emotion Prediction}
\vspace{-.1cm}
To preserve the original semantics of each node $v_i$, we employ a concatenation operation $\|$ to combine its hidden features from each layer $l$ in HDAGNN, resulting in the final node representation $H_{v_i}$.
Additionally, considering node consistency within a conversation turn $i$, we take the sum of the event representation $H_{e_i}$ and the state representation $H_{s_i}$ as the emotion representation $H_i$ of the utterance $u_i$, and input it into a fully connected layer to predict the emotion label:
\begin{equation}
\label{equ8}
H_i=\|_{l=0}^L (h_{e_i}^l+h_{s_i}^l),
\end{equation}
\begin{equation}
\label{equ9}
\hat{y_i} = \mathop{\operatorname{Softmax}}(W_z H_i + b_z).
\end{equation}

\vspace{-.15cm}
\section{Experiments}

\vspace{-.15cm}
\subsection{Setup}
\vspace{-.1cm}

\noindent
\textbf{Datesets:}
Our ESIHGNN is evaluated on four widely-used benchmark datasets: IEMOCAP~\cite{IEMOCAP}, MELD~\cite{MELD}, EmoryNLP~\cite{EmoryNLP}, and DailyDialog~\cite{DailyDialog}.
The dataset statistics are presented in Table~\ref{label_1}.
To assess performance, we apply the evaluation metrics utilized in previous studies~\cite{position-RGAT, Dialogxl}.
This entails calculating micro-averaged F1 for DailyDialog and weighted-average F1 for the other datasets.

\noindent
\textbf{Baselines:}
We compare our proposed ESIHGNN with state-of-the-art baseline methods, including
recurrence-based methods: DialogueRNN~\cite{DialogueRNN}, COSMIC~\cite{COSMIC}, DialogueCRN~\cite{DialogueCRN}, BiERU~\cite{BiERU}, MVN~\cite{MVN}, CoMPM~\cite{CoMPM}, and DialogueINAB~\cite{DialogueINAB};
and graph-based methods: DialogueGCN~\cite{DialogueGCN}, KET~\cite{KET}, KI-Net~\cite{KI-Net}, RGAT~\cite{position-RGAT}, SKAIG~\cite{SKAIG}, DAG-ERC~\cite{DAG-ERC}, CoG-BART~\cite{CoG-BART}, and MM-DFN~\cite{MM-DFN}.

\begin{table}[tb]
  \centering
  \vspace{-.3cm}
    \caption{Statistics of the datasets.
    ``A.~L.'' and ``A.~S.'' denote the average number of utterances and speakers per dialogue, respectively.
    }
  \scalebox{0.92}{
    \begin{tabular}{lccc}
      \toprule
      \multirow{2}[1]{*}{Datasets} & \# Dialogues & \# Utterances & \multirow{2}[1]{*}{A.~L./A.~S.}\\
      \cmidrule{2-2}\cmidrule(l){3-3} 
      & train / val / test & train / val / test &  \\
      \cmidrule{1-4}\morecmidrules\cmidrule{1-4}
      IEMOCAP & 100/20/31 & 4810/1000/1623 & 50/2 \\
      MELD & 1038/114/280 & 9989/1109/2610 & 10/2.7 \\
      EmoryNLP & 713/99/85 & 9934/1344/1328 & 14/3.5 \\
      DailyDialog& 11118/1000/1000 & 87170/8069/7740 & 8/2 \\
      \toprule
    \end{tabular}
  }
  \label{label_1}
\vspace{-.28cm}
\end{table}

\noindent
\textbf{Implementation Details:}
We use the PyTorch framework and the AdamW optimizer on two RTX 3090 GPUs for code implementation.
Our approach involves performing hyperparameter searches for the learning rate, dropout rate, batch size, and number of HDAGNN layers.
We set $\omega=1$ as the default setting for overall performance comparisons, but in Subsection \ref{sec: Ablation Study} we present ablation results for $\omega$ ranging from 1 to 3.
The initial dimensions for the node features and edge features are set to 1024 and 768, respectively.
These dimensions are consistent across all hidden layers, remaining at 300.
The reported results of our approach are the average of five test runs.

\vspace{-.15cm}
\subsection{Results and Analysis}
\vspace{-.05cm}
\label{sec:Result and Analysis}
The results of the four datasets are summarized in Table~\ref{label_2}.
It can be observed that:
(1) Our proposed ESIHGNN achieves state-of-the-art performance, except for competitive results in MELD.
This is probably attributed to the complex and ambiguous contexts present in short conversations with multiple speakers in MELD, which limit ESIHGNN's capability to effectively capture event-state interactions.
(2) Among recurrence-based methods, COSMIC outperforms DialogueRNN when external knowledge is incorporated, while our method surpasses DAG-ERC among graph-based methods.
These findings highlight the importance of incorporating external knowledge.
(3) Our ESIHGNN surpasses graph-based methods that do not consider future utterances on all datasets, confirming the superiority of the event-state connection structure and the effectiveness of ESIHGNN.
Although considering future utterances can enhance the model's understanding of the current context~\cite{BiERU,SKAIG}, it is not practical for real-time conversations.
(4) Graph-based methods consistently outperform recurrence-based methods on most CER datasets, indicating their superior capability to model conversational context.
Our method achieves better results compared to existing graph-based methods on IEMOCAP and EmoryNLP, showing its superiority in modeling conversation, particularly for longer conversations (with an average length of 50) in IEMOCAP.

\begin{table}[tbp]
  \centering
  \vspace{-.3cm}
  \caption{Results of our method and state-of-the-art baselines.
``\:\!$^\text{a}$" and ``\:\!$^\text{b}$" denote the incorporation of external information and the disclosure of future utterance information, respectively.
}
\scalebox{0.835}{
\begin{tabular}{{lcccc}}
  \toprule
Method & IEMOCAP & MELD&EmoryNLP& DailyDialog \\

\cmidrule{1-5}\morecmidrules\cmidrule{1-5}
 \multicolumn{3}{c}{~~~~~~~~\emph{Recurrence-based methods}}\\
  DialogueRNN & 64.76 & 63.61 &37.44&  57.32\\
  $\text{COSMIC}^\text{a, b}$ & 65.28 & $\textbf{65.21}$ & 38.11&58.48\\  
  $\text{DialogueCRN}^{\text{b}}$&66.20&58.39&-&-\\
  $\text{BiERU}^{\text{b}}$&65.22&60.84&-&-\\
    $\text{MVN}^{\text{b}}$&65.44&59.03&-&-\\
  $\text{CoMPM}^{\text{b}}$&65.79&64.62&37.44&59.63\\
    $\text{DialogueINAB}^{\text{b}}$&67.22&57.78&-&-\\
  
  \midrule
   \multicolumn{3}{c}{\emph{Graph-based methods}}\\
  DialogueGCN & 64.18 & 58.10 &-&-\\
    $\text{KET}^{\text{a}}$& 59.56 & 58.18 & 34.39&53.37\\
    $\text{KI-Net}^{\text{a}}$& 66.98 & 63.24 & -&57.30\\
    $\text{RGAT}^{\text{b}}$& 65.22 & 60.91 & 34.42&54.31\\
    $\text{SKAIG}^{\text{a, b}}$& 66.96 & 65.18 & 38.88&59.75\\
    DAG-ERC & 68.03 & 63.56 &39.02&59.33\\
    $\text{CoG-BART}^{\text{b}}$& 66.18 & 64.81 & 39.04&56.29\\
    $\text{MM-DFN}^{\text{b}}$ & 68.18 & 59.46 &-&-\\
    \midrule
$\textbf{ESIHGNN}^{\text{a}}$& $\textbf{68.53}$ & 63.92 &$\textbf{39.56}$&$\textbf{59.78}$\\
  \toprule
\end{tabular}}
  \label{label_2}
    \vspace{-.2cm}
\end{table}

\vspace{-.15cm}
\subsection{Ablation Studies}
\vspace{-.05cm}
\label{sec: Ablation Study}
In our ablation study, we analyze the contributions of different components in our ESIHGNN: edge construction, edge representation, IntraESI (Intra-turn Event-State Interactions), and window size $\omega$.
The edge construction module distinguishes our work from previous approaches like DAG-ERC~\cite{DAG-ERC} and SKAIG~\cite{SKAIG}, as it allows the analysis of relations at a coarser scale.
For example, $\texttt{-}\{event\texttt{-}to\texttt{-}event\}$ denotes the removal of the \texttt{xWant} and \texttt{oWant} relations.
To evaluate the effect of different edge representations, we use ``\:\!trainable\:\!'', which replaces all edge representations with trainable embeddings, and ``\:\!0/1'', which limits the relation types to the set $\mathcal{R}=\{0, 1\}$.
Table~\ref{label_3} presents the ablation results.
It can be seen that:
(1) Removing any of the coarse-grained relations in ESIHGNN leads to a decrease in overall performance, particularly in participants' reactions to events (i.e., $event\texttt{-}to\texttt{-}state$).
This suggests that emotion recognition is influenced by both previous events and speakers' emotional states.
(2) Removing knowledge-based edges results in a more significant performance decline compared to removing trainable embedding-based edges.
Additionally, both ``\:\!trainable\:\!'' and ``\:\!0/1'' show poor performance without encoding external knowledge in edge representations.
These findings suggest that external knowledge can enhance the information interaction between nodes.
(3) Ablating the IntraESI module noticeably decreases the performance of ESIHGNN, suggesting that there is an interplay between the speaker's events and emotional states.
(4) Increasing $\omega$ may not significantly affect the performance of ESIHGNN, indicating that the event-state interactions are localized.

\begin{table}[t]
\vspace{-.1cm}
  \centering
\caption{
Results of ablation studies on the four datasets.
}
\scalebox{0.83}{
\begin{tabular}{lcccc}
  \toprule
Method & IEMOCAP & MELD&EmoryNLP& DailyDialog\\
  \cmidrule{1-5}\morecmidrules\cmidrule{1-5}
  ESIHGNN                                     & 68.53    & 63.92 &39.56&59.78\\
  ESIHGNN~($\omega=2$)                        & 68.22 & 63.88 & 39.42&59.56\\
  ESIHGNN~($\omega=3$)                        & 68.17 & 63.91 &39.51&59.69\\
  \midrule

$\texttt{-}\{event\texttt{-}to\texttt{-}event\}$ & 67.70& 63.80 &39.38 & 59.52\\  
$\texttt{-}\{state\texttt{-}to\texttt{-}event\}$ & 68.03& 63.85 &39.50 & 59.68\\  
$\texttt{-}\{event\texttt{-}to\texttt{-}state\}$ & 67.56& 63.76 & 39.18& 59.22\\  
$\texttt{-}\{state\texttt{-}to\texttt{-}state\}$ & 67.60& 63.81 & 39.30& 59.47\\  
    \midrule
trainable                                        & 67.92 & 63.78& 39.40&59.48\\
0/1                                              &  67.68& 63.81& 39.37&59.50\\
\midrule
\texttt{-}IntraESI  &67.29 &  63.78& 39.05& 59.03\\  
  
  \toprule
\end{tabular}
}
  \label{label_3}
  \vspace{-.28cm}
\end{table}

\section{Conclusion}
\label{sce:conclusion}
This paper proposes ESIHGNN, a novel approach that incorporates the speaker's emotional state into conversational context modeling for conversational emotion recognition, based on a heterogeneous directed acyclic graph neural network.
The experimental results on four benchmark datasets demonstrate that our approach achieves competitive or state-of-the-art performance when compared with baselines.
Moreover, our ablation studies confirm the effectiveness of event-state interactions and emphasize the superiority of knowledge-enriched edge representations.
In future work, we plan to contribute a knowledge graph where emotion serves as the head entity, filling gaps in both state-to-state and state-to-event relations.

\newpage
\bibliographystyle{IEEEbib}
\bibliography{custom}

\end{document}